# A Hypervolume Based Approach to Rank Intuitionistic Fuzzy Sets and Its Extension to Multi-criteria Decision Making Under Uncertainty

K. Deveci and O. Guler

*Abstract*—**Ranking Intuitionistic Fuzzy Sets (IFS) with distance based methods requires to calculate the distance between IFS and a reference point which is known to have the maximum (positive ideal solution) or minimum (negative ideal solution) value. These group of approaches assume that as the distance of an IFS to the reference point decreases, the similarity of IFS with that point increases. This is a misconception, because an IFS which has the shortest distance to positive ideal solution does not have to be the furthest from negative ideal solution for all circumstances when the distance function is nonlinear. This paper gives a mathematical proof of why this assumption is not valid for any of the non-linear distance functions and suggests a hypervolume based ranking approach as an alternative to distance based ranking. In addition, the suggested ranking approach is extended as a new multi-criteria decision making method, HyperVolume based ASsessment (HVAS). HVAS is applied for multi-criteria assessment of Turkey's energy alternatives. Results are compared with three distance based multi-criteria decision making methods: TOPSIS, VIKOR, and CODAS.**

*Index Terms*—**Intuitionistic fuzzy set, distance function, hypervolume, multi-criteria decision making.**

## I. INTRODUCTION

FUZZY logic theory and ordinary fuzzy sets developed by Zadeh [1] has an important role in reasoning with uncertainty. It provides an effective way of modeling human thinking, perception, and decisions. Although fuzzy logic is an effective tool, the level of uncertainty it can describe is limited. In order to define uncertainty in a broader manner, ordinary fuzzy sets are extended to several new types of fuzzy sets such as Intuitionistic Fuzzy Sets (IFS) [2], pythagorean fuzzy sets [3], hesitant fuzzy sets [4], fuzzy multisets [5], neutrosophic sets [6], spherical fuzzy sets [7], type-2 fuzzy sets [8], etc. Among these, IFS is one of the most powerful extensions of ordinary fuzzy sets [9]–[14]. It uses membership and non-membership degrees to define a hesitant phenomenon, rather than a single membership degree that is used in ordinary fuzzy sets. As a result, IFS are regarded to be capable of modeling uncertainty in a more efficacious way than ordinary fuzzy sets.

Decision making is one of the active research fields where IFSs are frequently used to rank/select the most appropriate candidate over the set of available alternatives under uncertainty. In order to rank these alternatives, decision making methods in the literature follow different approaches which can be classified in four groups: outranking approach (Preference Ranking Organization Method for Enrichment of Evaluation and Elimination Et Choix Traduisant la Realite), pairwise comparison approach (Analytical Network Process, Analytical Hierarchy Process, Measuring Attractiveness by a Categorical Based Evaluation Technique), distance based approach (Technique for Order Preference by Similarity to Ideal Solution, Višekriterijumsko Kompromisno Rangiranje, Evaluation Based on Distance from Average Solution, Combinative Distance Based Assessment), and other approaches (Decision Making Trial and Evaluation Laboratory, Axiomatic Design, decision making with similarity measures, linear programming, aggregation operators, etc.). Among these approaches, distance-based decision making methods mostly considers ranking alternatives by using Euclidean or Hamming distance with respect to positive or negative ideal solution (or both). However, it is known that use of these measures with Intuitionistic Fuzzy (IF) extensions may result with non-robustness, inadmissibility, and indifference problems [15].

So how should IFSs be ranked with distance based approaches? Because of IFS's nature, one basic principle that should be followed during ranking is: IFV with greater membership and lower non-membership value should be given priority during ranking. However, there exists different situations, for example when both of the membership and non-membership values of one IF number is greater/lower than the other. To address this issue, several distance-based ranking measures have been proposed in the literature. To this end, Szmidt and Kacprzyk [16] suggested four distance measures based on Euclidean and Taxicab distances. Grzegorzewski [17] proposed a distance based on Hausdorff measure which is the generalization of Euclidean and Hamming distances. Wang and Xin [18] offered another distance measure by averaging the Hamming and Hausdorff distances. Ngan et al. [19] suggested a distance measure based on Hamming distance proposed by Szmidt and Kacprzyk [16] by evaluating the difference in maximum cross evaluation factor. Xu [20] has defined weighted distance measure to consider the importance of different sets in decision making. Mahanta and Panda [21] suggested another non-linear distance measure that performs better under high hesitancy compared to existing measures. Yang and Chiclana

This work was supported in part by Istanbul Technical University under Grant MDK-2021-43419. *(Corresponding author: K. Deveci).* K. Deveci also received partial support from TUBITAK-BIDEB.

K. Deveci is with the Energy, Science & Technology Department, Istanbul Technical University, Istanbul, 34469, Turkey, and also with Stuttgart-German University, 34820 Istanbul, Turkey (e-mail: deveci@itu.edu.tr).

O. Guler is with Energy, Science & Technology Department, Istanbul Technical University, Istanbul, 34469, Turkey(e-mail:onder.guler@itu.edu.tr).

[22] proposed a distance measure which extends the Hamming and Euclidean distances and considers IFS with 3 dimensions (including membership, non-membership, and hesitancy indices). Luo et al. [23] and Chen and Deng [24] suggested another distance measures by pointing the importance of using hesitancy as the third dimension in calculating distance of IFS. In these studies, researchers have evaluated membership and non-membership elements of IFS in a similar manner by assuming them to be the coordinates of a 2 or 3 dimensional space, and they have measured geometrical distance over these coordinates with respect to a reference point for ranking purposes. However, the information carried by non-membership element of an IFS is opposite of its' membership element and should not be treated similar to membership element in order to prevent IFS from any loss of information.

Later, researchers preferred to apply some novel modifications to known distance measures. To this end, Xiao [25] suggested a new distance measure which is based on Jensen-Shannon divergence. Hatzimichailidis et al. [26] proposed another distance measure where information of each IFS is stored in a matrix form. Cheng et al. [14] suggested another distance measure which is calculated by similarity matrix. Jiang et al. [27] introduced another distance measure which is based on the intersection of the diagonal of a square area and the transformed isosceles triangle of IFSs. Nguyen [28] suggested a knowledge-based distance function for IFS that considers the hesitancy as lack of information and calculates the degree of fuzziness and intuitionism at the same time. Although new approaches have added different perspectives to the ranking problem, the inadmissibility, non-robustness, and indifference problems of ranking IFS still persist. This reason, there is still a need for a new performance metric that can rank IFSs considering the problems mentioned above.

The contribution of this study, and what differentiates this work from the previously mentioned references are as follows:

- Most of the cited papers underline the problem of ranking IFS with distance measures and they suggest new nonlinear distance functions in which ranking of IFSs are claimed to be more accurate than other distance measures. We prove that any of the nonlinear distance measures cannot provide robust ranking of IFSs.
- To overcome this issue, a hypervolume (HV) based ranking method is suggested for IFSs.
- A new Multi-criteria Decision Making (MCDM) method, HyperVolume based ASsesment (HVAS), is introduced under uncertainty and an application is given.

The rest of the paper is as follows. Section II briefly describes some basic concepts related with IFS and hypervolume. Section III describes the problem of ranking IFSs with nonlinear distance functions. In Section IV, the suggested HV based ranking approach for IFSs is given with a case study. This section also introduces HVAS. A brief discussion is given in Section V and concluding remarks are given in Section VI.

## II. PRELIMINARIES

In this section, some preliminary concepts which are related with distance measures, IFS, and hypervolume will be reviewed.

### A. Intuitionistic Fuzzy Sets

Atanassov introduced the concept of IFS in order to extend the capabilities of ordinary fuzzy sets and explained the concept over an election metaphor in [29]. Let the percentage of electorates (k) is known for a selected government. Using ordinary fuzzy sets, the membership degree of electorates for the selected government becomes $\mu = k/100$, and the non-membership degree of the electorates which are not voted for government becomes $v = 1 - \mu$. Here, non-membership degree cannot be considered in more detail because it includes the degree of electorates who have voted against the government (opposition) and who have not voted at all. The idea of IFS born from this point: if the membership and non-membership degrees of electorates are defined in advance, the degree of electors who have not voted at all ($\pi$) can be defined as $\pi = 1 - \mu - v$. This reason, IFS contains more information and can better define uncertainty than ordinary fuzzy sets.

*Definition 1:* Let $X$ be a finite universe of discourse, an IFS $\tilde{A}$ in $X$ is defined as follows.

$$\tilde{A} = \{(x, \mu_{\tilde{A}}(x), v_{\tilde{A}}(x)) | x \in X\} \qquad (1)$$

where; $\mu_{\tilde{A}}(x),\ v_{\tilde{A}}(x) \in [0,1]$ and $\mu_{\tilde{A}}(x) + v_{\tilde{A}}(x) \leq 1$. The degree of hesitancy of $\tilde{A}$ ($\pi_{\tilde{A}}$) can be defined as follows.

$$\pi_{\tilde{A}} = 1 - \mu_{\tilde{A}}(x) - v_{\tilde{A}}(x) \qquad (2)$$

*Definition 2:* The score (S) [30] and accuracy (H) [31] functions of an IFS can be defined as follows.

$$S(\tilde{A}) = \mu_{\tilde{A}} - v_{\tilde{A}} \qquad (3)$$

$$H(\tilde{A}) = \mu_{\tilde{A}} + v_{\tilde{A}} \qquad (4)$$

As stated in [32], two IFS $\tilde{A}$ and $\tilde{B}$ can be compared by using score (S) and accuracy (H) functions as follows:

1. If $S(\tilde{A}) < S(\tilde{B})$, then $\tilde{A}$ is smaller than $\tilde{B}$ ($\tilde{A} < \tilde{B}$).
2. If $S(\tilde{A}) = S(\tilde{B})$, then:
   - If $H(\tilde{A}) = H(\tilde{B})$, then $\tilde{A}$ and $\tilde{b}$ represent the same fuzzy value ($\tilde{A} = \tilde{B}$).
   - If $H(\tilde{A}) < H(\tilde{b})$, then $\tilde{A}$ is smaller than $\tilde{b}$ ($\tilde{A} < \tilde{B}$).
   - If $H(\tilde{A}) > H(\tilde{b})$, then $\tilde{A}$ is greater than $\tilde{b}$ ($\tilde{A} > \tilde{B}$).

*Definition 3:* Let $\tilde{A}_l = (\mu_{\tilde{A}_l}, v_{\tilde{A}_l})$ and $\tilde{w}_l = (\mu_{\tilde{w}_l})$ where ($l$=1,2,3… $q$) be collections of intuitionistic and ordinary fuzzy sets, respectively. The Intuitionistic Fuzzy Arithmetic (IFA) aggregation operator is defined as in Eq. 5 [33]. In IFA aggregation operator, it is assumed that the sum of membership degrees of the weights is greater than zero ($\sum_{l=1}^{q} \mu_{\tilde{w}_l} > 0$). Otherwise, a sum of weights equal to 0 means that all DM evaluations are completely vague and new DMs should be chosen for evaluation.

$$IFA(\tilde{A}_1, \dots \tilde{A}_q; \tilde{w}_1, \dots \tilde{w}_q) = \left( \frac{\sum_{l=1}^{q} \mu_{\tilde{A}_l} \mu_{\tilde{w}_l}}{\sum_{l=1}^{q} \mu_{\tilde{w}_l}}, \frac{\sum_{l=1}^{q} v_{\tilde{A}_l} \mu_{\tilde{w}_l}}{\sum_{l=1}^{q} \mu_{\tilde{w}_l}} \right) \quad (5)$$

*Definition 4:* Let $d$ is a function which is defined in IF space and $\tilde{A}$, $\tilde{B}$ and $\tilde{C}$ are IFSs. Then, $d(\tilde{A}, \tilde{B})$ is called as a distance measure between $\tilde{A}$ and $\tilde{B}$ if the following conditions are satisfied.

Symmetry condition: $d(\tilde{A}, \tilde{B}) = d(\tilde{B}, \tilde{A})$.

Identity of indiscernibles: $d(\tilde{A}, \tilde{B}) = 0$, if and only if $\tilde{A} = \tilde{B}$.

Triangle inequality: $d(\tilde{A}, \tilde{B}) \leq d(\tilde{B}, \tilde{C}) + d(\tilde{A}, \tilde{C})$.

*Definition 5:* Let $\tilde{A} = \{x, \mu_{\tilde{A}}(x), \nu_{\tilde{A}}(x) | x \in X\}$ and $\tilde{B} = \{x, \mu_{\tilde{B}}(x), \nu_{\tilde{B}}(x) | x \in X\}$ be two IF numbers. Multiplication of two IFS is performed as follows.

$$\tilde{A} \otimes \tilde{B} = (x, \mu_{\tilde{A}}(x) . \mu_{\tilde{B}}(x),$$
$$\nu_{\tilde{A}}(x) + \nu_{\tilde{B}}(x) - \nu_{\tilde{A}}(x) . \nu_{\tilde{B}}(x) | x \in X) \quad (6)$$

*Definition 6:* The positive ideal solution ($\widetilde{PIS}$) is the greatest IF number and defined as $\widetilde{PIS} = (1,0)$. Similarly, the negative ideal solution ($\widetilde{NIS}$) is the smallest IF number and defined as $\widetilde{NIS} = (0,1)$.

### B. Hypervolume

Hypervolume (HV) can be defined as the volume of n-dimensional space. In multi-objective optimization, mainly with multi-objective metaheuristic algorithms, HV indicator is used to measure the quality of set of solutions with respect to a reference point by evaluating the solutions in terms of their closeness and divergence. It is used to compare the performance of algorithms over benchmark functions as a performance indicator [34], as a ranking method [35], and as a fitness assignment method [36]–[38]. HV indicator measures the size of the dominated space between the set of solution(s) and the reference point that also exists in $\mathbb{R}^n$ [39], [40].

*Definition 7:* For a given set of points $P \subset \mathbb{R}^n$ and a reference point $r \subset \mathbb{R}^n$, the HV indicator of $P$ which is the size of the dominated space above point $r$ can be defined as follows.

$$HV(P) = \wedge (\{q \in \mathbb{R}^n | \exists p \in P\}: p \leq q, q \leq r) \quad (7)$$

where $\wedge$ is the Lebesgue Measure. Note that HV indicator of a set of points $P$ can be defined as *length* in 1D space, *area* in 2D space, and *volume* in 3D space.

## III. PROBLEM STATEMENT

### A. Measuring Similarity with Non-linear Distance Measures

From the symmetry condition of distance measures (Definition 4), regardless of where the distance is measured from, distance between any of the two points will provide the same value. For example, $d(\widetilde{NIS}, \widetilde{PIS})$ will provide the same result with $d(\widetilde{PIS}, \widetilde{NIS})$ if function $d$ satisfies the symmetry condition. At this point, a question may arise: how to apply distance measures to rank IFS if a convenient comparison cannot be performed over two IF number due to the presence of symmetry condition? During ranking of IFS, researchers first chose a reference point whose value is known to be the most or least desired ($\widetilde{PIS}$ or $\widetilde{NIS}$). Next, the similarity of other IF numbers to that reference point are measured in terms of their distance to the reference point and IFS are ranked accordingly. In other words, ranking of IFSs with distance functions is possible if and only if a pivot reference point is chosen, only this way the value of distance function becomes meaningful for ranking purposes. Although it is not given enough importance, this approach requires to make a choice whether a positive or negative ideal solution is chosen as the reference point. The reason why it is not given enough importance is that, no matter the reference point is (whether $\widetilde{PIS}$ or $\widetilde{NIS}$), the idea of measuring the similarity of IF numbers over its' closeness to that reference point is thought to give meaningful and accurate results. But is it really?

*Definition 8:* A distance function $d(\Delta\mu, \Delta\nu)$ is called as *robust* if the ranking order of all IFS are preserved no matter $\widetilde{PIS}$ or $\widetilde{NIS}$ is selected as the reference point to measure the similarity.

In the current literature, linear distance functions are considered to be incapable of measuring the similarity of IFS for cases with high hesitancy and when the membership degree of an IF number is equal to its' non-membership degree. Most of the cited papers indicate that existing distance measures of IFS have problems such as indifference, non-robustness, inadmissibility and new non-linear distance measures are suggested to better rank the IFS (some examples can be found in [15], [25]). The main principle of ranking IFSs should give priority to IF numbers with greater membership and lower non-membership values. However, sometimes there occurs different cases those are hard to decide the order of ranking, when both membership and non-membership values are greater (or lower) than the other IF numbers.

*Theorem 1:* Any nonlinear IF distance function $d(\Delta\mu, \Delta\nu)$ cannot be *robust*.

*Proof:* Suppose that $x_1 = (\mu_1, \nu_1)$ and $x_2 = (\mu_1 + \Delta\mu, \nu_1 + \Delta\nu)$ are two unique intuitionistic fuzzy numbers where $\Delta\mu, \Delta\nu > 0$. Let $d(x, y)$ is a nonlinear distance function and assume that $x_1$ and $x_2$ have equal distance to $\widetilde{NIS} = (0,1)$: $d(\tilde{x}_1, \widetilde{NIS}) = d(\tilde{x}_2, \widetilde{NIS}) = l_1$. Let's assume that there exists a curve $c_1$ between $\tilde{x}_1$ and $\tilde{x}_2$, where any IF number on $c_1$ ($\tilde{x}_i$) have equal distance to $\widetilde{NIS}$: $d(\tilde{x}_i, \widetilde{NIS}) = (l_1)$, $i \in \{1,2,3, ...\}$. This indicates that any $\tilde{x}_i$ shows equal similarity.

If $d$ is *robust*, $x_1$ and $x_2$ must also lie on same curve (let's say, $c_2$) where all IF numbers on $c_2$ have equal distance to $\widetilde{PIS}$: $d(\tilde{x}_1, \widetilde{PIS}) = d(\tilde{x}_2, \widetilde{PIS}) = l_2$. Since $\widetilde{PIS} \neq \widetilde{NIS}$, $c_2$ and $c_1$ are non-concentric curves and cannot be identical ($c_1 \neq c_2$). Hence, depending on the order of non-linear distance function $d$, there will be only finite number of intersection points ($\tilde{x}_j$) of $c_1$ and $c_2$ which will be equidistant to $\widetilde{NIS}$ and $\widetilde{PIS}$. As a result, $\forall \tilde{x}_i \neq \tilde{x}_j$, ranking order of $\tilde{x}_i$ will be different than ranking order of $\widetilde{NIS}$. This reason, robustness will not be preserved for $d$ within the interval ($[\mu_1, \mu_1 + \Delta\mu]$, $[\nu_1, \nu_1 + \Delta\nu]$).

This theorem is visualized in Fig. 1 by using Euclidean distance within the defined interval ($[0,1], [0,1]$). Note that rather than intuitionistic values, l2 norm is used for demonstration purposes which means the abscissa represents x-coordinate and the ordinate represents y-coordinate regardless of the membership and non-membership functions ($x_1, x_2 \in \mathbb{R}^2$). Assume that $x_1 = (0.2, 0.6)$ and $x_2 = (0.4, 0.8)$, both $x_1$ and $x_2$ have the same Euclidean distance $d_E = 0.2$ to point $(0,1)$. Also assume that there exists a curve $c_1$ passing through $x_1$ and $x_2$, and any point on $c_1$ is equidistant to $(0,1)$. If

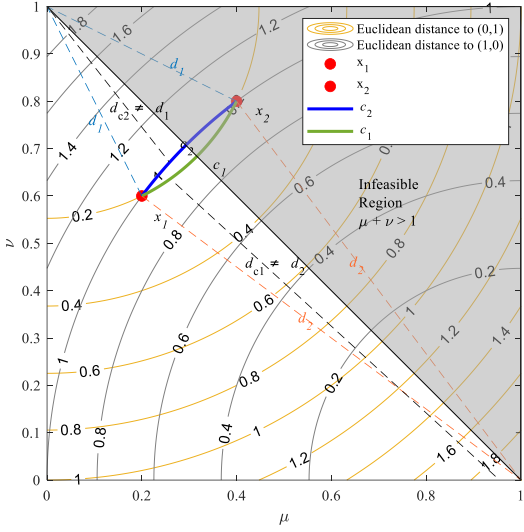

**Fig. 1.** Visualization of *Theorem 1*.

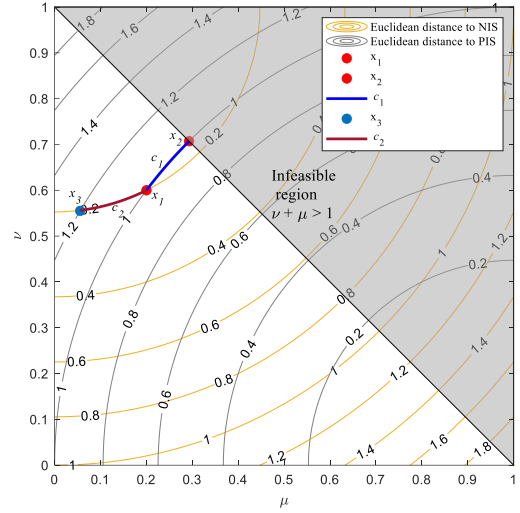

**Fig. 2.** Demonstration of the misconception of ranking IFS with non-linear distance functions.

Euclidean distance measure is *robust*, any point on $c_1$ must be equidistant to (1,0). However, any curve (namely, $c_2$) on which any point on it is equidistant from the point (1,0) can intersect $c_1$ at most two points (intersection points are $x_1$ and $x_2$ for the case given in Fig. 1). This means that $x_1$ and $x_2$ are equidistant to points (0,1) and (1,0); however, any of the remaining points on these curves are not. And this shows that Euclidean distance is not a *robust* measure. This is also the case when $l^2$ *norm* is changed with membership and non-membership values as represented with unshaded area in Fig. 1. Note that for Euclidean case, two arcs overlap if and only if they are concentric and their radii are same.

Let's discuss the robustness of distance measures with IF numbers under a different perspective by using Fig. 2. Suppose that two IF numbers, $\tilde{x}_1$ and $\tilde{x}_2$ are on curve $c_1$ so that any point on $c_1$ *is* equidistant to $\widetilde{PIS}$. Also suppose that another curve $c_2$ is passing through the points $\tilde{x}_1$ and $\tilde{x}_3$ which are all equidistant to $\widetilde{NIS}$. Using Aristotle's logic, it can be said that "*if $\tilde{x}_3$ is equal to $x_1$ and $\tilde{x}_1$ is equal to $\tilde{x}_2$, then, $\tilde{x}_3$ is equal to $\tilde{x}_2$*". Or mathematically, $[(\tilde{x}_3 \to \tilde{x}_1) \wedge (\tilde{x}_1 \to \tilde{x}_2)] \to (\tilde{x}_3 \to \tilde{x}_2)$.

In the case of Fig. 2, if the ranking of IF numbers $\tilde{x}_1$, $\tilde{x}_2$, and $\tilde{x}_3$ is performed with respect to $\widetilde{PIS}$, ranking will be found as $\tilde{x}_3 < \tilde{x}_2 = \tilde{x}_1$. If the ranking of the same IF numbers is performed with respect to $\widetilde{NIS}$, $\tilde{x}_3 = \tilde{x}_1 > \tilde{x}_2$ will be found. Please note that ranking with non-linear distance measures is performed by using a pivot point that is known to have the highest or lowest value such as $\widetilde{PIS}$ or $\widetilde{NIS}$, by using the idea of *"closer the distance measure, higher the similarity"*. As a result, the similarity of $\tilde{x}_3$ and $\tilde{x}_1$ with respect to $\widetilde{NIS}$ can be stated as "equal" and similarly, the similarity of $\tilde{x}_1$ and $\tilde{x}_2$ with respect to $\widetilde{PIS}$ can also be stated as "equal". Since this inference contradicts with previous ranking results, Euclidean distance cannot be said *robust*. Please note that this issue is not limited with Euclidean distance, as it is stated in *Theorem 1*, any non-linear distance measures will result with the same problem.

### B. Use of Distance Measures in MCDM Problems

Multi-criteria decision making (MCDM) is a set of methods that is used to rank the set of alternatives, or that is used to select the most appropriate alternative, by evaluating them under different set of criteria. Many of the MCDM problems include several conflicting criteria which are used to evaluate alternatives by using experts' opinions. Such decision making problems include human thinking and uncertainty, and therefore, most of the evaluation parameters cannot be defined precisely. At this point, fuzzy logic is referred to overcome the uncertainty in expert evaluations. Among fuzzy logic extensions, IFS are one of the widely preferred fuzzy sets in MCDM methods [14], [15].

MCDM methods can be grouped as pairwise comparison methods, outranking approaches, distance based methods, and other methods (i.e. optimization, similarity measures, etc.) which are applied to various fields such as assessment of energy systems [41], credit scoring analysis [42], selection of a suitable location [43], health [44], supplier selection [45], etc. In distance based crisp MCDM problems with $n$ criteria, each criterion of the alternatives represents one dimension in decision space $(c_A = (c_{1a}, c_{2a}, c_{3a}, \ldots c_{na}))$, and the distance of each alternative is calculated in $n$-dimensional decision space, between $c_A$ and the reference point $(R)$. However, the application of distance based MCDM methods with IFS is different: no matter how many criteria is included within the problem, distances of each alternative to reference points are calculated in 2-D space (membership and non-membership) for each criterion. Then, these distances are summed up to find the total distance of an alternative.

TABLE II
CALCULATED DISTANCE VALUES OF $\bar{X}$ WITH DIFFERENT MEASURES

| IFS | $P(\bar{X})$ | $N(\bar{X})$* | $d_{x,\widehat{PIS}}$ | $d_{x,\widehat{NIS}}$ | $d_{N,\widehat{PIS}}$ | $d_{N,\widehat{NIS}}$ | $d_{E,\widehat{PIS}}$ | $d_{E,\widehat{NIS}}$ | $d_{H,\widehat{PIS}}$ | $d_{H,\widehat{NIS}}$ | $HV_{net}$ |
|---|---|---|---|---|---|---|---|---|---|---|---|
| $\bar{X}_1$ | 0.3578 | 0.4913 | 0.5214 | 0.4175 | 0.7903 | 0.6423 | 0.4541 | 0.3597 | 0.5750 | 0.4250 | -0.36 |
| $\bar{X}_2$ | 0.4191 | 0.5683 | 0.4785 | 0.3310 | 0.6977 | 0.5304 | 0.4108 | 0.3053 | 0.5750 | 0.4250 | -0.42 |
| $\bar{X}_3$ | 0.4474 | 0.5466 | 0.4777 | 0.3389 | 0.6739 | 0.5788 | 0.3908 | 0.3206 | 0.5500 | 0.4500 | -0.29 |

*: $N(\bar{X})$ is obtained by using $\widehat{NIS}$ instead of $\widehat{PIS}$ in $P(\bar{X})$.

TABLE I
THREE IFS $\bar{X}_1$, $\bar{X}_2$, $\bar{X}_3$ FOR RANKING CASE

| |
|---|
| $\bar{X}_1 = \{(\tilde{x}_1, 0.2, 0.4), (\tilde{x}_2, 0.1, 0.2)\}$ |
| $\bar{X}_2 = \{(\tilde{x}_1, 0.3, 0.6), (\tilde{x}_2, 0.4, 0.4)\}$ |
| $\bar{X}_3 = \{(\tilde{x}_1, 0.2, 0.7), (\tilde{x}_2, 0.6, 0.3)\}$ |

In IF distance based MCDM methods, membership and non-membership values of IFS are used together as coordinates to synthesize information of similarity to an ideal point, in other words, similarity is calculated by comparing apples and oranges. For this reason, no matter how many new non-linear distance measures will be defined in the literature, inadmissibility, indifference, and non-robustness problems will remain as it is shown in *Theory 1*. In order to solve this issue, a new approach that can sort IFSs without using membership and non-membership values together as coordinates is needed.

## IV. SUGGESTED RANKING AND MCDM METHOD FOR IFS

As it is explained, use of nonlinear distance functions to rank IFS result with non-robustness issues and they are stuck in 2-D approaches no matter the number of elements in an IFS is. In addition, it is already discussed in the literature that linear functions cannot rank IFSs accurately when IFS has high hesitancy and equal membership/non-membership values. For these reasons, there is a need for a new metric that can clearly distinguish the meaning of membership and non-membership elements in an IFS during ranking and that can give robust and reliable ranking results.

### A. Ranking IFS with HV Indicator

In order to use HV indicator for ranking IFSs ($\bar{X}_i = \{\tilde{x}_{1i}, \tilde{x}_{2i}, \dots \tilde{x}_{ni}\}$), first decision spaces should be split and a reference point ($r$) needs to be determined. Unlike the previous approaches, 2 different decision spaces should be created for ranking purposes. The first decision space (*membership space, $U_\mu$*) includes the set of membership values ($U_\mu = \{\mu_1, \mu_2, \dots \mu_n\}$), whereas the second decision space (*non-membership space, $U_v$*) includes the set of non-membership values ($U_v = \{v_1, v_2, \dots v_n\}$) of IF numbers ($\tilde{x}_i$) in $\bar{X}$. The reason of separating the membership and non-membership spaces from each other is to prevent from any possible loss of information as explained with 2D distance based ranking approaches. Once the decision spaces are created, HV indicator is calculated for each IFS in $U_\mu$ and $U_v$ over $r$ by using Eq. 6. Then, the net HV is calculated by subtracting the HV of non-membership space from the HV of membership space ($HV_{net} = HV_\mu - HV_v$) and IFSs are ranked based on their $HV_{net}$ values.

TABLE III
RANKING OF IFSS BASED ON DISTANCE MEASURES AND
$HV_{NET}$ GIVEN IN TABLE II.

| Method | Ranking Order |
|---|---|
| $P(\bar{X})$ | $\bar{X}_3 \succ \bar{X}_2 \succ \bar{X}_1$ |
| $N(\bar{X})$ | $\bar{X}_2 \succ \bar{X}_3 \succ \bar{X}_1$ |
| $d_{x,\widehat{PIS}}$ | $\bar{X}_3 \succ \bar{X}_2 \succ \bar{X}_1$ |
| $d_{x,\widehat{NIS}}$ | $\bar{X}_1 \succ \bar{X}_3 \succ \bar{X}_2$ |
| $d_{N,\widehat{PIS}}$ | $\bar{X}_3 \succ \bar{X}_2 \succ \bar{X}_1$ |
| $d_{N,\widehat{NIS}}$ | $\bar{X}_1 \succ \bar{X}_3 \succ \bar{X}_2$ |
| $d_{E,\widehat{PIS}}$ | $\bar{X}_3 \succ \bar{X}_2 \succ \bar{X}_1$ |
| $d_{E,\widehat{NIS}}$ | $\bar{X}_1 \succ \bar{X}_3 \succ \bar{X}_2$ |
| $d_{H,\widehat{PIS}}$ | $\bar{X}_3 \succ \bar{X}_2 = \bar{X}_1$ |
| $d_{H,\widehat{NIS}}$ | $\bar{X}_3 \succ \bar{X}_2 = \bar{X}_1$ |
| $HV$ | $\bar{X}_3 \succ \bar{X}_1 \succ \bar{X}_2$ |

Suppose that three different IFS which are given in Table I are to be ranked by using P value [15], $d_x$ value [25], $d_n$ value [14], Euclidean distance, Hamming distance, and hypervolume metric. The distance of IFSs are calculated with respect to $\widehat{PIS}$ and $\widehat{NIS}$, and given in Table II. For HV calculations, a reference point of (-1, -1) is chosen for both $U_\mu$ and $U_v$, and greater $HV_{net}$ is given priority in ranking. When the corresponding nonlinear distance measures are applied to rank IFSs over $\widehat{PIS}$ and $\widehat{NIS}$, the order of ranking changes with respect to reference point taken (either $\widehat{PIS}$ or $\widehat{NIS}$) as it is given in Table III. However, only Hamming distance which is the only linear measure used in this test provided robust results regardless of the chosen ideal point.

### B. A new HV Based IF-MCDM method and its application

For decision making under uncertainty with multiple conflicting criteria, a new IF-MCDM method, HyperVolume based ASsesment (HVAS) is suggested as follows.

Step 1: Obtain Decision Maker (DM) evaluations on alternatives ($\tilde{A}_{j,i}^l$), importance of each criterion ($\tilde{w}_j^l$), and DM expertise for each criterion ($\tilde{e}_j^l$) where $l$ ($l \in \{1,2,\dots,q\}$) corresponds to the number of DM, $i$ ($i \in \{1,2,\dots n\}$) represents the number of alternatives, and $j$ ($j \in \{1,2,\dots m\}$) represents the number of criteria. Same notations also will be used in the below equations.

Step 2: Aggregate DM evaluations on alternatives ($\tilde{A}_{j,i}^l$) with DM expertise defined for each criterion ($\tilde{e}_j^l$) by using IFA given in Eq. 5 and obtain the aggregated evaluation matrix ($\tilde{A}_{ji}$).

Step 3: Aggregate DM evaluations on criteria importance ($\tilde{w}_j^l$) with DM expertise for each criterion ($\tilde{e}_j^l$) by using IFA given in Eq. 5 and obtain the matrix of aggregated criteria importance ($\tilde{w}_{ji}$).

Step 4: Normalize the aggregated evaluation matrix ($\tilde{A}_{ji}$) by using Eq. 8 [46] as follows.

$$\tilde{A}_{ji} = [\mu_{ji}, \nu_{ji}]$$
$$\tilde{N}_{ji} = \begin{cases} [\mu_{ji}, \nu_{ji}], & \text{if } j \in B \\ [\nu_{ji}, \mu_{ji}], & \text{if } j \in C \end{cases} \quad (8)$$

here $B$ and $C$ represent the set of benefit and cost criteria, respectively.

Step 5: Calculate the weighted normalized evaluation matrix ($\tilde{r}_{ji} = \tilde{N}_{ji} \otimes \tilde{w}_{ji}$) by using Eq. 6.

Step 6: Using weighted normalized evaluation matrix ($\tilde{r}_{ji}$) generate decision spaces. Then, calculate the HV metric for each decision space ($HV_\mu$ and $HV_\nu$) with respect to reference point ($r_j$) and obtain $HV_{net}$ ($HV_{net} = HV_\mu - HV_\nu$).

Step 7: Rank the alternatives based on the $HV_{net}$ values.

For comparison and testing purposes, a MCDM study for ranking energy alternatives in Turkey has been performed with HVAS and results are compared with three existing distance based IF MCDM methods in the literature: VIKOR [47], TOPSIS [48], and CODAS (modified from interval valued intuitionistic fuzzy CODAS in [33]).

### C. Experimental Setup

The steps of each MCDM method are summarized over a flowchart given in Fig. 3. In order to make an accurate comparison, same aggregation method given in Eq. 5 is used in all MCDM methods. Similarly, score and accuracy functions given in *Definition 2* is used to find the positive solution ($\widetilde{PS}$) and negative solution ($\widetilde{NS}$) values in all methods. $\widetilde{PIS}$ and $\widetilde{NIS}$ (1,0) and (0,1) could not be used because proposed methods include normalization steps that is performed by the best and worst alternatives in the available set. The distance functions remained same as the way they were suggested in articles. For CODAS, parameter $\tau$ is selected as 0.02 which is the suggested value in [49]. The parameter $v$ in VIKOR is selected as 0.5. For reproduction purposes, all MCDM codes are generated in MATLAB and will be publicly shared in Github (link will be shared).

3 DMs were evaluated a total of 17 criteria, 7 alternatives, and their own expertise with respect to each set of criteria. The available alternatives are selected as: biomass (A1), geothermal (A2), hydroelectric (A3), wind (A4), solar PV (A5), natural gas (A6), and coal (A7). During evaluation, DMs are allowed to use either their own IF evaluations or linguistic labels. The IF correspondence of linguistic evaluations which are used to evaluate alternatives and criteria importance are given Table A1 and Table A2, respectively. The selected set of criteria is modified from [50] and listed in Table A3 in supplementary file.

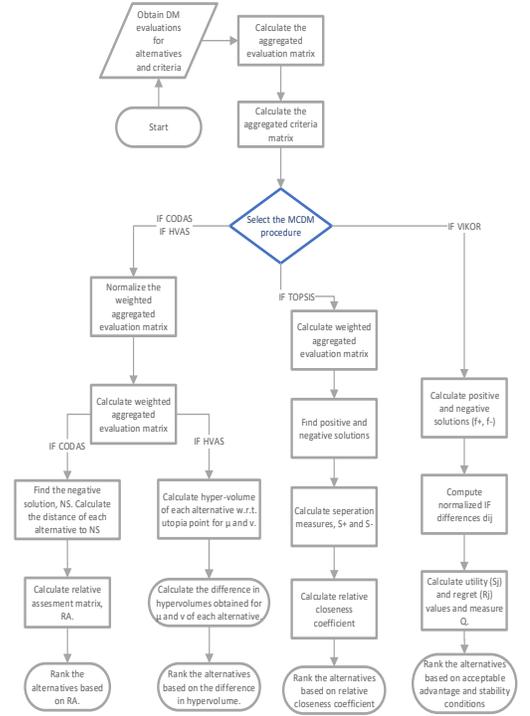

**Fig. 3.** Algorithm flowchart of MCDM methods.

TABLE IV
ORDER OF RANKING WITH RESPECT TO EACH
MCDM METHOD

| Method | Ranking Order |
|--------|---------------|
| TOPSIS | $\tilde{A}_5 \succ \tilde{A}_4 \succ \tilde{A}_3 \succ \tilde{A}_6 \succ \tilde{A}_1 \succ \tilde{A}_2 \succ \tilde{A}_7$ |
| VIKOR | $\tilde{A}_5 \succ \tilde{A}_3 \succ \tilde{A}_4 \succ \tilde{A}_1 \succ \tilde{A}_6 \succ \tilde{A}_2 \succ \tilde{A}_7$ |
| CODAS | $\tilde{A}_5 \succ \tilde{A}_4 \succ \tilde{A}_6 \succ \tilde{A}_2 \succ \tilde{A}_3 \succ \tilde{A}_1 \succ \tilde{A}_7$ |
| HVAS | $\tilde{A}_5 \succ \tilde{A}_4 \succ \tilde{A}_6 \succ \tilde{A}_2 \succ \tilde{A}_1 \succ \tilde{A}_3 \succ \tilde{A}_7$ |

### D. Results

The order of ranking obtained by selected MCDM methods are given in Table IV. From Table IV, one can see that TOPSIS, VIKOR, CODAS, and HVAS provide ranking orders with slight differences. Despite this slight difference, the best and worst alternatives (which are solar PV and coal) were the same in all of these methods. Although these results are problem specific, it can be said that there is a general correlation in the results with small variations. However, in some fields, these variations can be crucial (i.e., decisions relating to human health).

The proposed method (HVAS) provided very close results with CODAS method. This is probably because the use of Euclidean and Hamming distances in CODAS to measure the similarity to the ideal solution provides more reliable results than using only Euclidean or any other nonlinear measures. It is worth noting once again that the similarity in ranking orders of different MCDM methods will differ based on the MCDM problem and the expert evaluations.

## V. Discussion

When IFS are ranked by using distance based approaches, an ideal point is selected and the distance of each IFS is calculated with respect to the ideal point. The ranking process is performed based on the assumption that the IFS with shorter distance is more similar to the selected ideal point. In general, it is expected that this assumption will give the same ranking order regardless of whether the ideal point is $\widehat{PIS}$ or $\widehat{NIS}$. In other words, an IFS which has shortest distance to $\widehat{NIS}$ should be furthest from the $\widehat{PIS}$. However, it has been shown and proven that this assumption is not true for nonlinear distance measures. As a result, we believe that the reliability of any nonlinear IF distance measures, which are widely used in the literature, should be questioned. In addition, use of linear distance functions were also criticized in the literature [14], [15], [21]. This reason, it was thought that a new method could be used to rank IFSs instead of distance functions. To this end, hypervolume metric which is widely used in multi-objective optimization, is applied to rank IFS.

In distance based IF ranking methods, membership and non-membership values of IFSs are used as coordinates which leads to loss of information. In addition, it is known that the distance based ranking approaches cause inadmissibility, indifference, non-robustness issues. To prevent this, membership and non-membership spaces are separated from each other and their HV values are calculated over a reference point. Reference point selection is an important step of calculating HV values. First of all, the reference point must be selected outside of the decision space, otherwise HV measure cannot be calculated. Second, choosing a reference point very close or far away from decision space may result with inappropriate results. Let's recall the definition of HV indicator, it measures the dominated space between two or more points. The reference point should be selected in such a way that the dominated space at $U_\mu$ should be maximized at $\mu_j = 1$, and minimized at $\mu_j = 0$, where $j \in \{1,2,\ldots m\}$. Similarly, the dominated space at $U_\upsilon$ should be maximized at $\upsilon_j = 1$, and minimized at $\upsilon_j = 0$. For these reasons, $r_j \leq 0$. In addition, HV contribution at point $\mu_j, \upsilon_j = 0$ should be identity for both of the decision spaces. To this end, the reference point is chosen at $(-1,-1,\ldots)_{1 \times j}$ for each dimension of $j$ so that there will be no HV contribution at $\mu, \upsilon = 0$. In this way, the difference between the value of each element of an IFS and the value of corresponding reference point is guaranteed to be greater or equal to 1 ($\mu_j - \left(-r_j\right) \geq 1, \upsilon_j - \left(-r_j\right) \geq 1$). Otherwise, HV contribution of corresponding dimension would undesirably reduce the value of $HV_\mu$ (or $HV_\upsilon$). In the end, greater the dominated spaces of membership ($HV_\mu$) and smaller the dominated space of non-membership ($HV_\upsilon$) points out a greater IFS.

During MCDM process, the fact that HVAS gives very similar results as CODAS in selection of energy alternatives, may make the reader question whether it is possible to solve the non-robustness problem by hybridizing nonlinear distance functions with linear ones. In this case, it is thought that reaching any conclusions with at least one incorrect premise (in this case, measuring similarity to ideal solution with nonlinear distance functions) may result with *proton pseudos[1]*. Therefore, we believe that the use of distance measures is not reliable enough to rank IFSs. Also, HVAS allows DM to measure performance metric in single step for all criteria as in crisp MCDM methods, whereas distance based IF-MCDM methods calculate the distance for each criteria and then summing them up. This is because HVAS creates two separate decision spaces (namely, membership space $U_\mu$ and non-membership space $U_\upsilon$) and evaluates alternatives accordingly.

The problem with linear distance measures is that when membership and non-membership degrees are equal, indifference problem occurs, which is stated as counterintuitive [51]. However, it is believed that this issue is related to how hesitancy is processed with IFSs. In some cases, decision maker would be neutral to hesitancy, hence the effect of hesitancy would be neutral in decision making as well. Note that it is possible to add a hesitancy term to linear distance functions with a coefficient that states how prone DM to uncertain cases. However, in this case, the new function cannot provide *robustness* which is defined in *Definition 8* and also cannot provide symmetry condition of distance measures. Here is another advantage of using HV indicator to rank IFSs: it is possible to extend $HV_{net}$ metric by including the DM's reaction to hesitancy as in Eq. 9. The term $\alpha$ in Eq. 9 is named as "factor of perception" that evaluates the perception of DM to hesitancy and can take any value between [-1,1]. When DM is averse to hesitancy, $\alpha$ takes any value between (0,1], or if DM is prone to uncertainty, $\alpha$ takes any value between [-1,0). Lastly it is worth mentioning that although HVAS is proposed to give decisions under IF environment, it can be used with other fuzzy extensions, even with crisp decision sets.

$$HV_{net} = HV_\mu - HV_\upsilon - \alpha HV_\pi \qquad (9)$$

## VI. Conclusions

The distance based IF ranking methods rank IFS by using the distance between IFS and any selected ideal point. These approaches assume that "shorter the distance, higher the similarity to the selected ideal point". In this study, it has been shown that among these approaches, use of nonlinear distance measures is not reliable for ranking IFS because the ranking orders are changing based on the selected ideal solution. Whether it is fuzzy logic or classical logic, the method used must be robust and accurate when giving a decision. This reason, the need for a new metric to rank IFSs is obvious. Hence, in this study, a hypervolume based metric which is widely used in multi-objective optimization literature is used to rank IFSs. In addition, a new hypervolume based multi-criteria decision making method under uncertainty is also suggested and an application is given. Results are compared with well-known distance based VIKOR, TOPSIS, and CODAS methods.

As future research, HVAS applications with crisp and fuzzy extensions will be studied for different MCDM problems.

---

[1]: In Aristotle's logic, *proton pseudos* means "the fundamental misconception". It expresses the fact that if the premise in an inference is false, even though the approach is performed correctly, false statements will be reached (i.e., rare things are valuable and blind horses are rare. Then, blind horses are valuable).